\documentclass[UTF8]{llncs}
\usepackage{mathrsfs}
\usepackage{amssymb}
\usepackage{amsmath}
\usepackage{mathptmx}
\usepackage{graphicx}
\usepackage{verbatim}
\usepackage{subfigure}
\usepackage{cite}
\usepackage{amsfonts}
\usepackage{url}
\usepackage{array,tabularx}
\usepackage[misc]{ifsym}
\usepackage{multirow}
\usepackage{booktabs}
\begin{document}

\title{Knowledge Graph Embedding Bi-Vector Models for Symmetric Relation}
\author{Jinkui Yao$^1$ \and Lianghua Xu$^2$}
\institute{}
\footnotetext[1]{J. Yao (\Letter) \\
Jiangnan Institute of Computing Technology, WuXi, 214000, China\\
e-mail: pelops.yao@gmail.com}
\footnotetext[2]{L. Xu \\
Jiangnan Institute of Computing Technology, WuXi, 214000, China\\
}


\maketitle
\begin{abstract}

Knowledge graph embedding (KGE) models have been proposed to improve the performance of knowledge graph reasoning. 
However, there is a general phenomenon in most of KGEs, as the training progresses, the symmetric relations tend to zero vector, if the symmetric triples ratio is high enough in the dataset.
This phenomenon causes subsequent tasks, e.g. link prediction etc., of symmetric relations to fail.
The root cause of the problem is that KGEs do not utilize the semantic information of symmetric relations.
We propose KGE bi-vector models, which represent the symmetric relations as vector pair, significantly increasing the processing capability of the symmetry relations.
We generate the benchmark datasets based on FB15k and WN18 by completing the symmetric relation triples to verify models.
The experiment results of our models clearly affirm the effectiveness and superiority of our models against baseline.

\end{abstract}

\begin{keywords}
knowledge graph embedding, symmetry relation, bi-vector models
\end{keywords}

\section{Introduction}

The knowledge graph, a structured knowledge base, represents world's truth in a form that computer can easily process.
As the basis of question answering and knowledge inference, etc., the knowledge graph has received extensive attention from academia and industry.

In recent years, knowledge graph reasoning has made significant progress. There are two main branches, logical reasoning and representation learning, each with its own advantages and disadvantages.
Logical reasoning based on the rigorous mathematical foundation is difficult to solve the computational bottleneck of the combinatorial explosion.
Knowledge representation learning based on statistics has attracted more attention because of the development of machine learning and deep learning at present, but it is limited by the incompleteness and the scale of the knowledge base.

Usually, each fact of the knowledge graph is represented by a triple $(h, r, t)$, where $h$ and $t$ are the head entity and the tail entity, respectively, and $r$ is the relation between them.
\begin{figure}\label{Fig_1}
\centering
\includegraphics[width=.95\columnwidth]{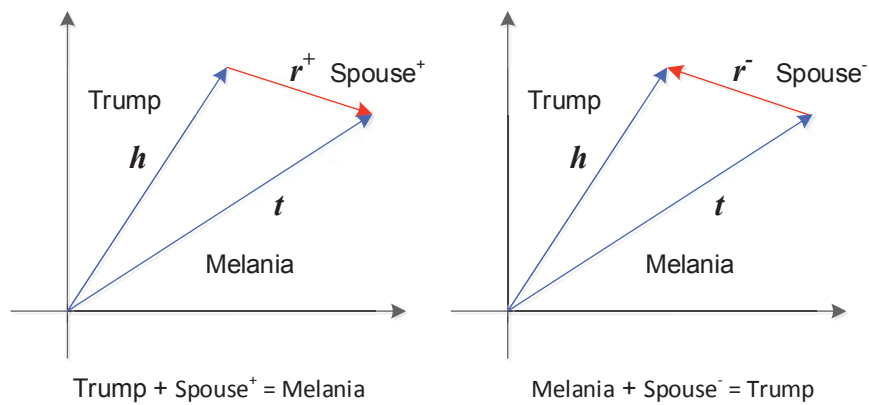}\\
\caption{\hspace*{\fill}symmetric relations spouse\hspace*{\fill}}
\end{figure}

For example, the triple $(Trump,\ spouse,\ Melania)$ means that Trump's spouse is Melania, in which Trump is the head entity, the spouse is the relation, and Melania is the tail entity.
Semantically, relation $spouse$ is symmetric, shown in figure\ref{Fig_1}, $(Trump,\ spouse,\ Melania)$ and $(Melania,\ spouse,\ Trump)$ simultaneously hold.

KGE aims to embed the entities and relations into low-dimensional real vectors, and then learns the representations of them.
TransE \cite{Bordes2013} is the earliest KGE model and has derived a series of models called Trans series models or Trans models.
Most of Trans models based on vector addition calculation, which are difficult to apply well in symmetric relations.

We propose bi-vector models extended the Trans models for symmetric relations.
Different from the Trans models using a single vector to represent the entity or relation, We adopt bi-vector to represent symmetric relation.
The score functions of the two subvectors are calculated separately.
With the increase of training epochs, the two subvectors are separated step by step.
And then, models can distinguish the two directions of the symmetric relation

Two benchmark datasets, FB15k-SYM and WN18-SYM construced by us for running bi-vector models on them
The experimental results show that our method can effectively improve the triple prediction accuracy of symmetry relations.
The main contributions of this paper as follow.
\begin{enumerate}
\item We propose bi-vector models which improve the prediction accuracy of symmetric relations.
\item The symmetric semantic information of relations is combined with KGE, which is a new research method of knowledge graph reasoning.
\item We run the model on the extended benchmark datasets and verify the effectiveness and advantages of the models.
\end{enumerate}

\section{Related Works}

We extend three popular KGE models, TransE, TransH \cite{Wang2014} and TransD \cite{Ji2015}, using bi-vector.
Therefore, we firstly introduce these.

\subsection{TransE, TransH and TransD}

\paragraph{•TransE}, the first KGE model proposed, regards relation $ r $ as the translation from entity $ h $ to $ t $.
Entity $ t $ should be in the nearest neighborhood of $ h+r $. The score function is defined as
\begin{equation}\label{scorefunctranse}
f_r(h,t)=\big\|h+r-t\big\|_{L_n}.
\end{equation}
Where $L_n$ is usually as $L_1$ norm or $L_2$ norm.
TransE can slove 1-1 relations effectively, but it is not suitable for handling 1-n, n-1 and n-n relations.

\paragraph{•TransH}
projects entities $h$ and $t$ into the hyperplane which relation $r$ located.
TransH calculates $h_{\bot}=h-\omega_r^{\top}h\omega_r$, $t_{\bot}=t-\omega_r^{\top}t\omega_r$ before calculating score function,
\begin{equation}
f_r(h,t)=\big\|(h-\omega_r^{\top}h\omega_r)+r-(t-\omega_r^{\top}t\omega_r)\big\|_{L_n}.
\end{equation}
Where $L_n$ is usually as $L_2$ norm.
TransH is more accurate than TransE in terms of recognition rate of 1-n, n-1 and n-n relations.

\paragraph{•TransD}
believes that combinations of entities and relations can distinguish the relation more finely.
The combination of entity $h$ and relation $r$ correspondences association matrix $M_rh$.
The calculation of score function uses the product of entity and association matrix, form as $h_{\bot}=M_{rh} h$, $t_{\bot}=M_{rh} t$. The score function is defined as
\begin{equation}
f_r(h,t)=\big\|M_{rh} h+r-M_{rh} t\big\|_{L_n}.
\end{equation}
Where $L_n$ is usually as $L_2$ norm.

\subsection{Other Models}
\paragraph{•Translation based methods}. In addition to TransE(H,D) that we have already mentioned, translation based methods cover the following models.
\textbf{TransR} \cite{Lin2015} build entity and relation embedding independent spaces, in which, entities $h,t \in \mathbb{R}^k$, and relation $r \in \mathbb{R}^d$. A projection matrix $M_r\in \mathbb{R}^{k\times d}$ has been set, and the score funcion is defined as $f_r(h,t)=\big\|M_rh+r-M_rt\big\|_2^2.$
\textbf{TransSparse} \cite{Ji2016} set two separate relation sparse matrices $M_r^h(\theta _r^h)$ and $M_r^t(\theta _r^t)$ to deal with the issue of sparse data. The score function is defined as $\big\|M_r^h(\theta _r^h)h+r-M_r^t(\theta _r^t)t\big\|_{L_n}$.
\textbf{TransF}  reduces the cost of calculation of relation projection by modeling subspaces of projection matrices, and the score function is defined as $f_r(h,t))=\big\|(\sum_{i=1}^s\alpha _r^{(i)}U^{(i)} + I)h + r + (\sum_{i=1}^s\beta _r^{(i)}V^{(i)} + I)t\big\|_{L_n}$, where $s\in\mathbb{R}$,${U^{(i)}},{V^{(i)}} \in\mathbb{R}^{d_e \times d_r}$, $\alpha _r^{(i)}U^{(i)}$ and $\beta _r^{(i)}V^{(i)} + I)$are the corresponding coefficients of ${U^{(i)}}$ and ${V^{(i)}}$.

\paragraph{•Tensor based methods}.
\textbf{DistMult} \cite{Yang2015} adopts a relation-specific diagonal matrix $M_r$ to represents the characteristics of a relation.
The score function $f_r(h,t)=hM_rt$ is a bilinear function, which score of positive triples should be higher than negative triples.
\textbf{HolE} \cite{Nickel2016} employs circular correlations by holographic to create compositional representations, and has advantages of computation efficiency and representing scalability.
\textbf{RESCAL} \cite{Nickel2011} adopt tensor factorization to estimate relation axis.\textbf{ComplEX} \cite{Welbl2016} embed the entities and relation to complex space, then computes loss vaule.
\paragraph{•Other related methods}.
\textbf{SE}\cite{Bordes2012b} defines two relation-specific matrices for $h,t$, i.e. $M_{r,1},M_{r,2}$, and defines the score function as $f_r(h,t)=\big\|M_{h,r}h-M_{t,r}t\big\|_1$.
There are many other KGE models try to try to use various embedding methods, such as \textbf{Neural Tensor Network (NTN)}\cite{Socher2013} , \textbf{Semantic Matching Energy (SME)}\cite{Bordes2012}, \textbf{SLM}, \textbf{TransA}, \textbf{lppTransD}, etc.

However, these works did not utilize the semantic information of relations properties.
We believe that the semantic information of the relations properties are of value and can improve the performance of the KGE models.

\section{Methodology}

In order to overcome the lack of support for symmetric relations in KGE, we made the following efforts.
First of all we describe the defects of Trans models in handling symmetric relations, and analyze the causes of it.
Then, we propose three new models that extends the Trans models to improve the performance of handling symmetry relations in KGE, which are named TransE-SYM, TransH-SYM and TransD-SYM.
Finally, we give the definition of the loss functions for these models.

\subsection{Problems and causes}\label{problems}
Knowledge graph can be represented as a set of ordered triples of entities and relations.
Each triple in Knowledge graph is essentially a binary relation, which have the properties of symmetry, anti-symmetric, reflexive, anti-reflexive and transitive properties.
This paper focuses on the relation's properties of symmetry.
In graph, symmetric relation have two directed edges in opposite directions.

KGE represents each relation, including symmetric relation, as a low-dimensional real vector.
However, a single vector cannot represent two opposite directions.

We take TransE as an example to illustrate the problem of symmetric relations.
TransE learns the embedding feature from equation $h + r = t$ when triplets $(h, r, t)$ holds.
TransE's scoring function is defined as $f_r(h, t)=\big\|h + r - t\big\|_{L_n}$. When the function $f_r(h,t) = 0$, it means $h + r = t$.

Assuming that there is a symmetric relation $r_s$ and triple $(h, r_s, t)$ in $KG$, then $h + r_s = t$, ie $r_s = t - h$.
Since $r_s$ is symmetric, then the symmetric triple $(t, r_s, h)$ should hold too, satisfying $t + r_s = h$, ie $r_s = h - t$.

Obviously, if both $r_s = t - h$ and $r_s = h - t$ are correct, if and only if $r_s$ is an \emph{additive identity} of vector, ie $r_s \equiv \vec{0}$, the conclusion contradicts with the conditions of TransE model.

Taking the symmetric relation $spouse$ as an example, shown in figure\ref{Fig_1}.
When the fact $(Trump, spouse, Melania)$ holds, the fact $(Melania, spouse, Trump)$ holds too.
let $e_{Trump}$, $e_{Melania}$ and $e_{Melania}$ denote entities \emph{Melania, Trump} and relation \emph{spouse}, respectively. Then,
\begin{equation}\label{trump1}
e_{Trump} + r_{spouse} = e_{Melania}
\end{equation}
\begin{equation}\label{trump2}
e_{Melania} + r_{spouse} = e_{Trump}
\end{equation}
let Equation(\ref{trump1}) + Equation(\ref{trump2}),
\[e_{Trump} + e_{Melania} + 2r_{spouse} = e_{Melania} + e_{Trump}\]
we have
\begin{equation}\label{trump3}
r_{spouse} = 0
\end{equation}

According to the KGE preset, the relations $r_{spouse}$ should be a non-zero real vector, and Equation (\ref{trump3}) contradicts with the condition.
The root cause of the above problem is that the symmetric relation is represented by single vector, and the single vector cannot express semantic bifurcation of symmetric relation.

\begin{figure}
\centering
\includegraphics[width=.45\columnwidth]{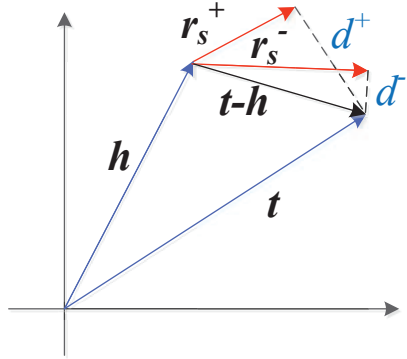}\\
\caption{\hspace*{\fill}
The two subvectors of the symmetric relation $r_s$ are denoted as $r^+_s$ and $r^-_s$, and the results of their score functions can be denoted as distances $d^+$ and $d^-$. The distance of $d^-$ is shorter, that is, the value of $f_{r_s}(h,t)$ is smaller, and $r^-_s$ is selected as the training subvector.
\hspace*{\fill}}
\label{Fig_2}
\end{figure}

\subsection{Our Method}

Aiming at these problems, bi-vector models for symmetric relation are presented in this study.

\emph{Knowledge graph} $KG$, $KG{}=\{(h,r,t)\} \subseteq E \times R \times E$, Where $E$ and $R$ are entities set and relations set, respectively.

\emph{Symmetric relation} $r_s$ , if $h$ and $t$ are entities of knowledge graph $KG$, $r_s$ is the relation of  $KG$, and $(h, r_s, t)\subseteq KG$, $(t, r_s, h)\subseteq KG$, then relation $r_s$ is symmetric relation.

Different from most of KGE models, which represent entities and relations as single vector, we represent the symmetric relation $r _ s $ as a bi-vector with two subvectors, $r^+_s$ and $r^-_s $.
Then, in each epoch of learning, the score functions of the two subvectors are calculated, and the better score is selected as the current result.
Let $f_{r_{s}}(h,t)$ be the score function of the Trans series model, as show in Equation(\ref{minfunc})
\begin{equation}\label{minfunc}
f_{r_{s}}(h, t) = \min(f_{r^+_s}(h, t),f_{r^-_s}(h, t))
\end{equation}

We have extended three different Trans models, which differ in their respective score functions.
In TransE, score function is $f_{r_s}(h, t)=\big\|h + r_s - t\big\|_{L_n}$, where ${L_n}$ is L1 norm or L2 norm, and the score functions of subvectors are shown as Equation array(\ref{equ1}),
\begin{align}\label{equ1}
\left\{\begin{array}{c}
f_{r^+_s}(h, t)=\big\|h + r^+_s - t\big\|_{L_n}\\
f_{r^-_s}(h, t)=\big\|h + r^-_s - t\big\|_{L_n}
\end{array}\right.
\end{align}
$f_{r_{s}}(h,t)$ and $r_s$ should be substituted into the following loss function,
\begin{equation}\label{loss}
L=\sum_{(h,r_s,t)\in \Pi _{r_{s}}} \sum_{(h',r_s,t')\in \Pi' _{r_{s}}} [\lambda + f_{r_{s}}(h,t)-f_{r_{s}}(h',t')]_{+}.
\end{equation}
where $\lambda > 0$ denotes the margin of hyperplane, and $[x]_+$ denotes $\max(x,0)$.
Similarly, the score function of the TransH model is shown in Equation array (\ref{scoreTransH}).
\begin{align}\label{scoreTransH}
\left\{\begin{array}{l}
f_{r^+_s}(h, t)=\big\|(h-\omega_{r^+_s}^\top h \omega_{r^+_s}) + r^+_s - (t-\omega_{r^+_s}^\top t \omega_{r^+_s})\big\|_{L_n}\\
f_{r^-_s}(h, t)=\big\|(h-\omega_{r^-_s}^\top h \omega_{r^-_s}) + r^-_s - (t-\omega_{r^-_s}^\top t \omega_{r^-_s})\big\|_{L_n}\\
f_{r_{s}}(h, t) = \min(f_{r^+_s}(h, t),f_{r^-_s}(h, t))
\end{array}\right.
\end{align}
The score function of the TransH model is shown in Equation array (\ref{scoreTransD}).
\begin{align}\label{scoreTransD}
\left\{\begin{array}{l}
f_{r^+_s}(h, t)=\big\|M_{r^+_sh} h + r^+_s - M_{r^+_st} t\big\|_{L_n}\\
f_{r^-_s}(h, t)=\big\|M_{r^-_sh} h + r^-_s - M_{r^-_st} t\big\|_{L_n}\\
f_{r_{s}}(h, t) = \min(f_{r^+_s}(h, t),f_{r^-_s}(h, t))
\end{array}\right.
\end{align}
The loss functions of them are calculated according to Equation (\ref{loss}).

\section{Experiments and results}

\begin{table}
\begin{center}
\caption{\textbf{Statistics of several popular datasets.}$|\mathcal{E}|$ is the number of entity, and $|\mathcal{R}|$ is the number of relations, train/test/valid is the number of train/test/valid set.
$ SYM^{\ddagger} $is number of symmetric triple,
$ SYM^{\dagger} $is number of complement symmetric triple,
$ ALL $is number of triple in dataset,
$ ALL + SYM^{\dagger} $is number of triple in dataset after complement,
$ \frac{SYM^{\ddagger}}{ALL} @\%$ is percentage of symmetric triples in the train/test/valid set,
$ \frac{SYM^{\ddagger}+2SYM^{\dagger}}{ALL+SYM^{\dagger}} @\%$is percentage of symmetric triples in the train/test/valid set after complement.
}\label{dataset}
\begin{tabular}{l@{\quad}rllll}
\toprule
\multicolumn{1}{l}{\rule{0pt}{12pt}
Dataset}
&\multicolumn{1}{c}{$|\mathcal{E}|$}
&\multicolumn{1}{c}{$|\mathcal{R}|$}
&\multicolumn{1}{c}{train/test/valid}
&\multicolumn{1}{c}{$ \frac{SYM^{\ddagger}}{ALL} @\%$}
&\multicolumn{1}{c}{$ \frac{SYM^{\ddagger}+2SYM^{\dagger}}{ALL+SYM^{\dagger}} @\%$}\\[2pt]
\midrule
FB15k & 14,951 & 1,345 & 483,142/50,000/59,071 & 7.15/0.94/0.744 & 8.69/8.41/8.34 \\
FB15k-237 & 14,541 & 237 & 272,115/17,535/20,466 & 12.48/1.44/1.13 & 14.97/2.65/2.58\\
FB13 & 75,043 & 13 & 316,232/5,908/23,733 & 1.31/0.00/0.00 & 1.42/0.00/0.00\\
WN18 & 40,943 & 18 & 141,442/5,000/5,000 & 20.97/0.52/0.72 & 22.38/19.07/19.01\\
WN11 & 38,696 & 11 & 112,581/2,609/10,544 & 1.41/0.06/0.00 & 1.54/0.20/0.08 \\
WN18RR & 40,943 & 11 & 86,835/3,134/3,034 & 34.15/0.83/1.19 & 36.05/27.38/27.98 \\
\bottomrule
\end{tabular}
\end{center}
\end{table}

\subsection{Dataset analysis and preprocessing}\label{preprocessing}

In this study, we compared and analyzed the commonly used knowledge graph embedding benchmark data sets FB15k, FB15k-237, FB13, WN18, WN11 and WN18RR.
FB15k, FB15k-237 and FB13 are extracted from Freebase\cite{Bollacker2008}, which is a large-scale common sense knowledge base provided the general facts of the world.
Freebase was acquired by Google and is still under maintenance.
WN18, WN11 and WN18R aextract from WordNet \cite{Miller1994} and provide semantic knowledge of words.

We count the ratio of the symmetric relations in the data set shown in the table \ref{dataset}.
It can be seen that the proportion of symmetric data of the WN18 and FB15k data set are relatively high.

The proportion of symmetric data for relation $r$ is denoted as $\zeta_r$ by the paper.
We regard $r$ as symmetric relation When $\zeta_r$ exceeds the threshold\footnote{In this paper, the threshold is set to 0.5.}.

As shown in the table\ref{dataset-SYM}, in WN18, the relation $\_verb\_group$ has 1139 triples, of which 1060 are symmetric triples, and the ratio of symmetric triples is about 0.93.
Semantically, the relation $\_verb\_group$ is the meaning of verb grouping, which is obviously a symmetric relation.
From the perspective of data distribution, the symmetry rate of the relation $\_verb\_group $ is 0.93, and we believe it is symmetrical.

In order to simplify the problem, in this paper, symmetry is only judged by data distribution.We complement the missing symmetric triples in dataset of the symmetric relation.
A more formal description is, if relation $r_s$ in knowledge graph $KG$ is symmetric, for $\forall (h,r_s,t)\in KG $, if $(t,r_s,h)\notin KG $ and then $ KG = KG \cup (t,r,h)$.

\begin{table}
\begin{center}
\caption{\textbf{Symmetric relation examples in FB15k and WN18.} SYM is the number of symmetric relations, ALL is number of relations and $\frac{SYM}{ALL}$ is the proportion of the symmetric relations in the total number of relations.}
\label{dataset-SYM}
\begin{tabular}{lllll}
\toprule
Dataset&\multicolumn{1}{c}{\rule{0pt}{12pt}Relation}&SYM&ALL&$\frac{SYM}{ALL}$\\
\midrule
\multirow{5}{*}{FB15k}&
/military/military\_combatant/force\_deployments/.../combatant & 78 & 84 & 0.929 \\
~&/base/fight/crime\_type/p.../crime/criminal\_conviction/guilty\_of & 20 & 21 & 0.952 \\
~&/base/twinnedtowns/twinned\_town/.../town\_twinning/twinned\_towns & 20 & 21 & 0.952 \\
~&/base/contractbridge/.../bridge\_tournament\_standings/second\_place & 18 & 19 & 0.947 \\
~&/sports/sports\_position/.../sports-\_team\_roster/position & 108 & 127 & 0.850 \\
\midrule
\multirow{4}{*}{WN18}&
\_derivationally\_related\_form & 27694 &29716& 0.931\\
~&\_verb\_group & 1060 &1139 & 0.931\\
~&\_similar\_to & 74 &81 & 0.914\\
~&\_also\_see & 830 &1300 & 0.638\\
\bottomrule
\end{tabular}
\end{center}
\paragraph{Remark}
\end{table}

\subsection{Benchmarks}
In order to show the superiority of our models, we compare the following benchmark KGE models.
\paragraph{•TransE} is the most widely used KGE model, also the earliest proposed KGE model.
\paragraph{•TransH} projects h and t to the hyperplane where r located, to solve the relations of 1-n, n-1 and n-n.
\paragraph{•TransD} uses the entity-relation matrix to obtain a more fine-grained distinction of realtion.

\subsection{Verification problem}
In order to verify the problem of the Trans models described in Section \ref{problems}, We have designed the following experiments, the steps are as follows.
\begin{enumerate}
\item \textbf{Training Trans models.} We train the TransE, TransH and TransD models on the datasets which are completed symmetric triples in Section \ref{preprocessing}.
\item \textbf{Constructing test dataset.}We randomly selected symmetric relations and entities in FB15k and WN18 to construct test sets. Each test set contains 10,000 symmetric triples named \textsl{FB15k-test-circle} and \textsl{WN18-test-circle}. The form of triples in test sets is $(e, r_s, e)$, where $r_s$ and $e$ are respectively symmetric relation and any entity. The triple example is as follows,\\$(05451384,\_derivationally\_related\_form,05451384)$,\\ $(04958634,\_verb\_group,04958634)$.
\item \textbf{Experimental results.}
According to Section \ref{problems}, if the symmetric triple is true, the relation tends to zero.
We run the test sets on models and the experimental results are shown in Table \ref{result_circle}.
Almost all randomly generated triples is true.
These models completely fail in dealing with \textbf{all of} symmetric relations.
\end{enumerate}

\begin{table}
\centering
\caption{Circle triple test result.}\label{result_circle}
\begin{tabular}{llll llll}
\toprule
Model&Train Dataset&Test Dataset&MR& MRR& H10& H3& H1\\
\midrule
TransE & FB15k-SYM & FB15k-test-circle & 1.000 & 1.000 & 1.000 & 1.000 & 1.000\\
TransH & FB15k-SYM & FB15k-test-circle & 1.000 & 1.000 & 1.000 & 1.000 & 1.000\\
TransD & FB15k-SYM & FB15k-test-circle & 1.000 & 1.000 & 1.000 & 1.000 & 1.000\\
TransE & WN18-SYM & WN18-test-circle & 1.000 & 1.000 & 1.000 & 1.000 & 1.000\\
TransH & WN18-SYM & WN18-test-circle & 1.000 & 1.000 & 1.000 & 1.000 & 1.000\\
TransD & WN18-SYM & WN18-test-circle & 1.000 & 1.000 & 1.000 & 1.000 & 1.000\\
\bottomrule
\end{tabular}{}
\end{table}
\subsection{Result of Experiment.}
Three bi-vector Trans models named TransE-SYM, TransH-SYM and TransD-SYM proposed by us.
Experimental code implementation reference open source project OpenKE\cite{han2018openke}.
These models run on datasets completed symmetric relation and get good results.
The experimental results are shown in Table \ref{result}.
Bi-vector models are superior to the original model in indicators of the link prediction task.

\begin{table}
\centering
\caption{\textbf{Experimental result}}\label{result}%
\begin{tabular}{l lllll lllll}
\toprule
\multirow{2}{*}{}&
\multicolumn{5}{c}{FB15k-SYM}&
\multicolumn{5}{c}{WN18-SYM} \\
\cmidrule(lr){2-6}\cmidrule(lr){7-11}
& MR& MRR& H10& H3& H1& MR& MRR& H10& H3& H1\\
\midrule
TransE & 66 & 0.490 & 0.683 & 0.461 & 0.206 & 493 & 0.371 & 0.711 & 0.544& 0.087\\
TransE-SYM & 51 & 0.534 & 0.772 & 0.598 & 0.329 & 467 & 0.485 & 0.836 & 0.705& 0.246\\
\midrule
TransH & 80 & 0.380 & 0.747 & 0.539 & 0.162 & 688 & 0.426 & 0.926 & 0.828 &0.026\\
TransH-SYM & 49 & 0.432 & 0.784 & 0.612 & 0.344 & 601 & 0.577 & 0.931 & 0.845 &0.120\\
\midrule
TransD & 185 & 0.265 & 0.519 & 0.297 & 0.148 & 711 & 0.416 & 0.928 & 0.787 & 0.145\\
TransD-SYM & 72 & 0.642 & 0.774 & 0.543 & 0.335 & 210 & 0.886 & 0.941 & 0.866 & 0.374\\
\bottomrule
\end{tabular}{}
\end{table}

\section{Conclusion}
This paper introduces symmetry semantics into KGE models, and points out the defect of the state-of-the-art KGE models learning symmetric relations.
Bi-vector models proposed by us can improve the situation of low recognition rate of symmetric relations in Trans models.

\end{document}